\documentclass[12pt]{article}
\usepackage{wrapfig}
\usepackage{graphicx}

\usepackage{times}

\newenvironment{sciabstract}{%
\begin{quote} \bf}
{\end{quote}}

\title{ A novel cluster internal evaluation index based on hyper-balls }

\author
{Jiang Xie,$^{1}$  Pengfei Zhao,$^{2}$ Shuyin~Xia,$^{3\ast}$ Guoyin Wang,$^{4}$ Dongdong Cheng$^{5}$\\
\\
\normalsize{$^{1}$College of Computer Science and Technology, Chongqing University of Posts and Telecommunications,}\\
\normalsize{Chongwen Road, Nan'an District Chongqing}\\
\normalsize{$^{2}$P. R. China, 400065.}\\
\\
}

\date{}

\begin{document} 


\baselineskip24pt


\maketitle


\begin{sciabstract}
It is crucial to evaluate the quality and determine the optimal number of clusters in cluster analysis. In this paper, the multi-granularity characterization of the data set is carried out to obtain the hyper-balls. The cluster internal evaluation index based on hyper-balls(HCVI) is defined. Moreover, a general method for determining the optimal number of clusters based on HCVI is proposed. The proposed methods can evaluate the clustering results produced by the several classic methods and determine the optimal cluster number for data sets containing noises and clusters with arbitrary shapes. The experimental results on synthetic and real data sets indicate that the new index outperforms existing ones.
\end{sciabstract}

\section{The process of generating hyper-ball}

\begin{figure}[hbpt!]
	\centering
	{\includegraphics[width = 0.9\textwidth]{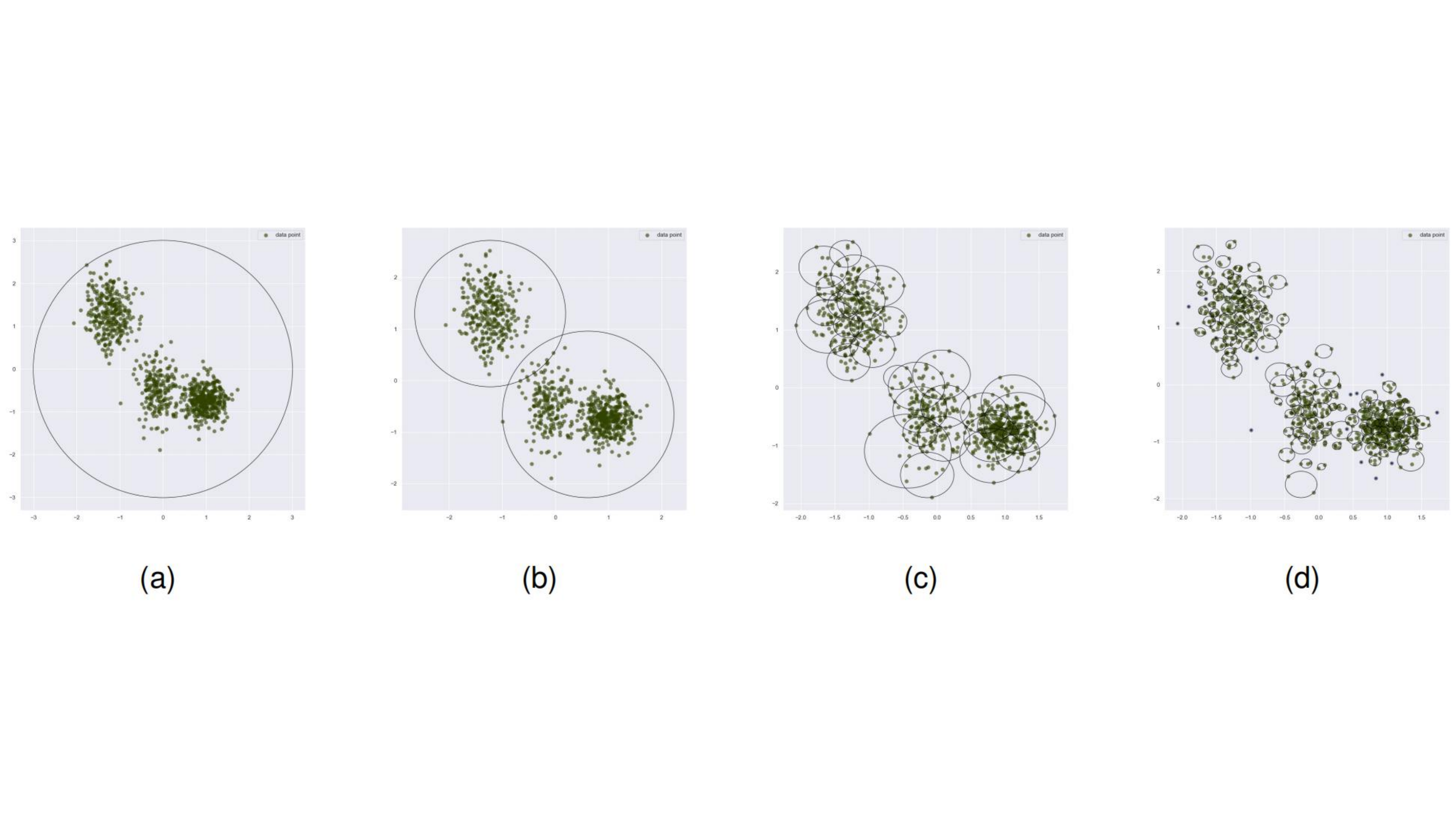}}
	\caption{The splitting process of hyper-balls on the dataset core vs nocore. And the blue dots indicate that there is only one sample in the hyper-ball.}
	\label{fig_sim}
\end{figure}

The inputs to most of the existing clusters are points or pixels, inspired by the GCHB\cite{01}, the multi-granularity characterization of the data set is carried out to obtain the hyper-balls. We have made a effective improvement in the process of generating hyper-ball(HB). The whole data is regarded as the coarse granularity, and then it is characterized by hyper-ball which is split and refined from coarse granularity to fine granularity, to realize a scalable, and robust computing process. Our approach is based on a basic assumption that the data with similar distribution form a hyper-ball. 

\textbf{Definition 1.} Given a data set D$\in{R^d}$, for each HB, $c_i$ is the center of gravity of all data points in HB, and $r_i$ is the maximum distance from all points $p_i$ in HB to  $c_i$, and $r_a$ is the average distance from all points $p_i$ in HB to $c_i$. The center c and radius r are defiend as follows:
\begin{equation}
\label{deqn_ex1a}
c_i =1/n \sum_{i=1}^{n}p_i.
\end{equation}
\begin{equation}
\label{deqn_ex2a}
r_i =max(||p_i-c_i||).
\end{equation}
\begin{equation}
\label{deqn_ex3a}
r_a =avg(||p_i-c_i||).
\end{equation}
Where $\left||.\right||$ denotes the 2-norm and n is the number of data points in HB.

The split way of GCHB is that if the value of the sub hyper-balls’ Distribution Measure(DM) is greater than that of the hyperball, it will be split, where DM is measured by computing the ration of the number data point $n_i$ and the sum radius $s_i$ in HB. However, this splitting method leads to too many hyper-ball, so we propose a more effective improvement method to determine the optimal cluster number for data sets containing noises and clusters with arbitrary shapes. 

\textbf{Definition 2.} Given a data set D$\in{R^d}$, for each HB, the balance degree BD is the absolute value of the difference between $r_i$ and $r_a$. Obviously, when the value of the balance degree is smaller, the data in the hyper-ball is more dense, that is to say, the effect of clustering is better at this time. Thus, BD is defined as follows: 
\begin{equation}
\label{deqn_ex4a}
BD =r_i-r_a.
\end{equation}

As shown in Fig.1, it is the splitting process of the hyper-balls. The kmeans clustering algorithm is more suitable for splitting hyper-ball since it has almost linear time complexity. And the main steps are designed as follows: Firstly, we use 2-means to divide the entire dataset into two sub-datasets which namely two sub-hyper-ball and then the balance degree of the two hyper-balls are computed separately. The hyper-ball will be divided iteratively until its balance degree becomes lower than the balance degree threshold and then the iteration will be stopped.
\section{The cluster internal evaluation index based on hyper-ball}
The cluster internal evaluation index uses the intracluster compactness to represent the similarity of the samples in the cluster and the intercluster separation to represent the distinct of the samples in different clusters, which aim to evaluate the clustering results and obtain the optimal clustering numer. As shown in Fig.2(a), $C_1$ represents one of the clusters in which there are four hyper-balls $HB_{1}$, $HB_{2}$, $HB_{3}$ and $HB_{4}$. The black dots represent the object points assigned to the hyper-ball, and the red triangles represent the center of the hyper-ball. Obviously, the distance between the two hyper-balls of $HB_{1}$ and $HB_{3}$ which calculated by the distance between the centers of the two hyper-balls minus the corresponding radius and indicated by the yellow line in the figure is the maximum distance between all hyper-balls in the cluster and we use this distance as the intracluster compactness of $C_1$ cluster. It is clear that $HB_{2}$ and $HB_{3}$ belong to the overlapping relationship, and we take the minimum distance between all non-overlapping hyper balls within a cluster in all clusters as the distance of overlapping hyper balls. If a cluster has only one hyperball, the same approach will be taken by us.

\begin{figure}[hbpt!]
	\centering
	{\includegraphics[width = 0.9\textwidth]{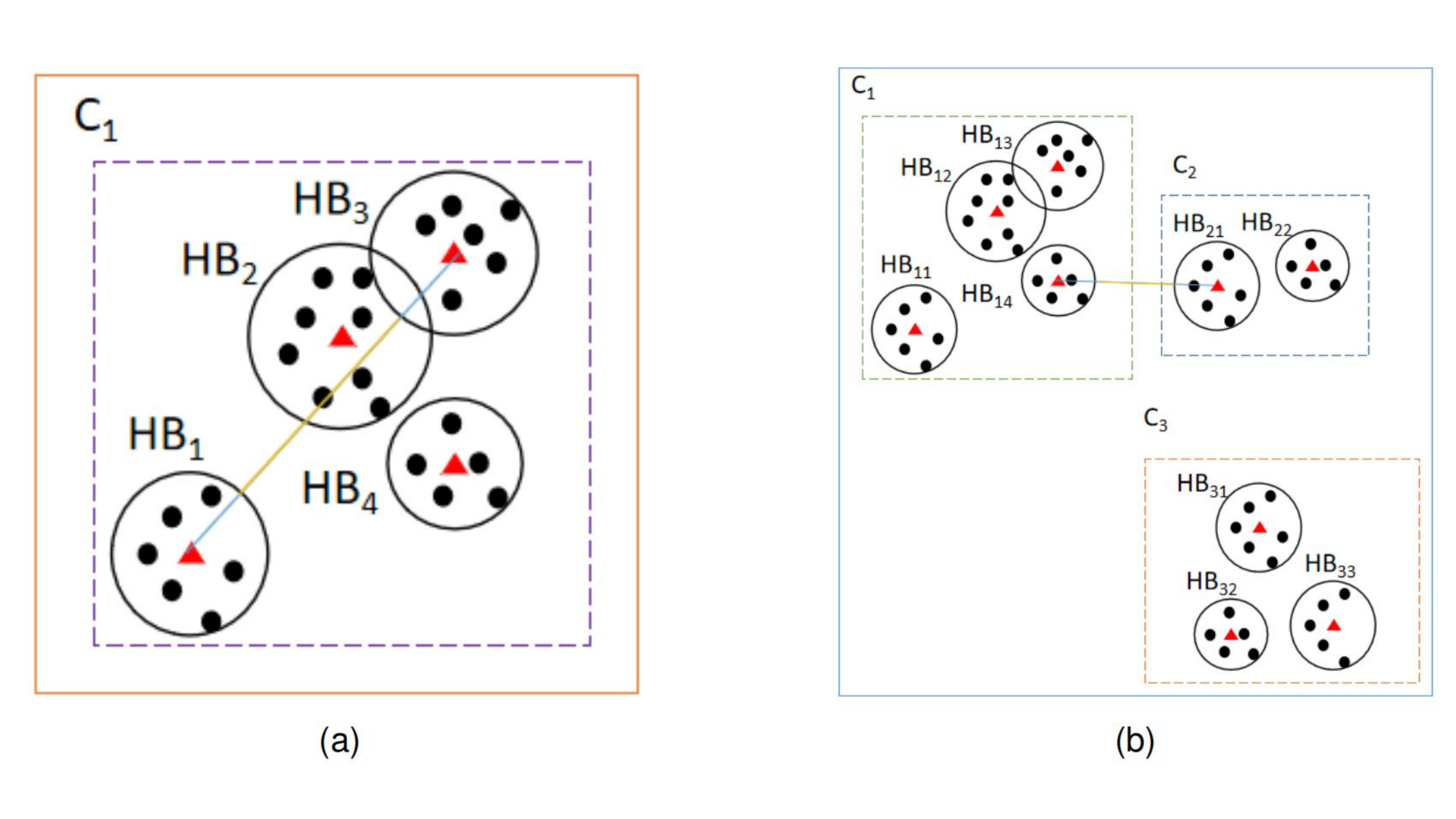}}
	\caption{(a) indicates how to compute intracluster compactness within a cluster.(b) shows how the intercluster separation between clusters is calculated.}
	\label{fig_sim1}
\end{figure}

As shown in Fig.2(b), we assume that the datasets have three clusters $C_1$, $C_2$, $C_3$ and now we will take the intercluster separation of $C_1$ as an example. $C_1$ represents the clusters in which there are four hyper-balls $HB_{11}$, $HB_{12}$, $HB_{13}$ and $HB_{14}$; $C_2$ represents the clusters in which there are two hyper-balls $HB_{21}$, $HB_{22}$; $C_3$ represents the clusters in which there are three hyper-balls $HB_{31}$, $HB_{32}$, $HB_{33}$. Obviously, the distance between the two hyper-balls of $HB_{14}$ and $HB_{21}$ is the minimum distances between the hyper-balls in this cluster and all hyper-ball in other clusters and we use this distance as the intercluster separation of $C_1$ cluster. And we define the distance as the smallest distance among all pairs of non-overlapping hyper-balls in different clusters if two hyper-balls in different clusters overlap each other.

\begin{figure}[hbpt!]
	\centering
	{\includegraphics[width = 0.9\textwidth]{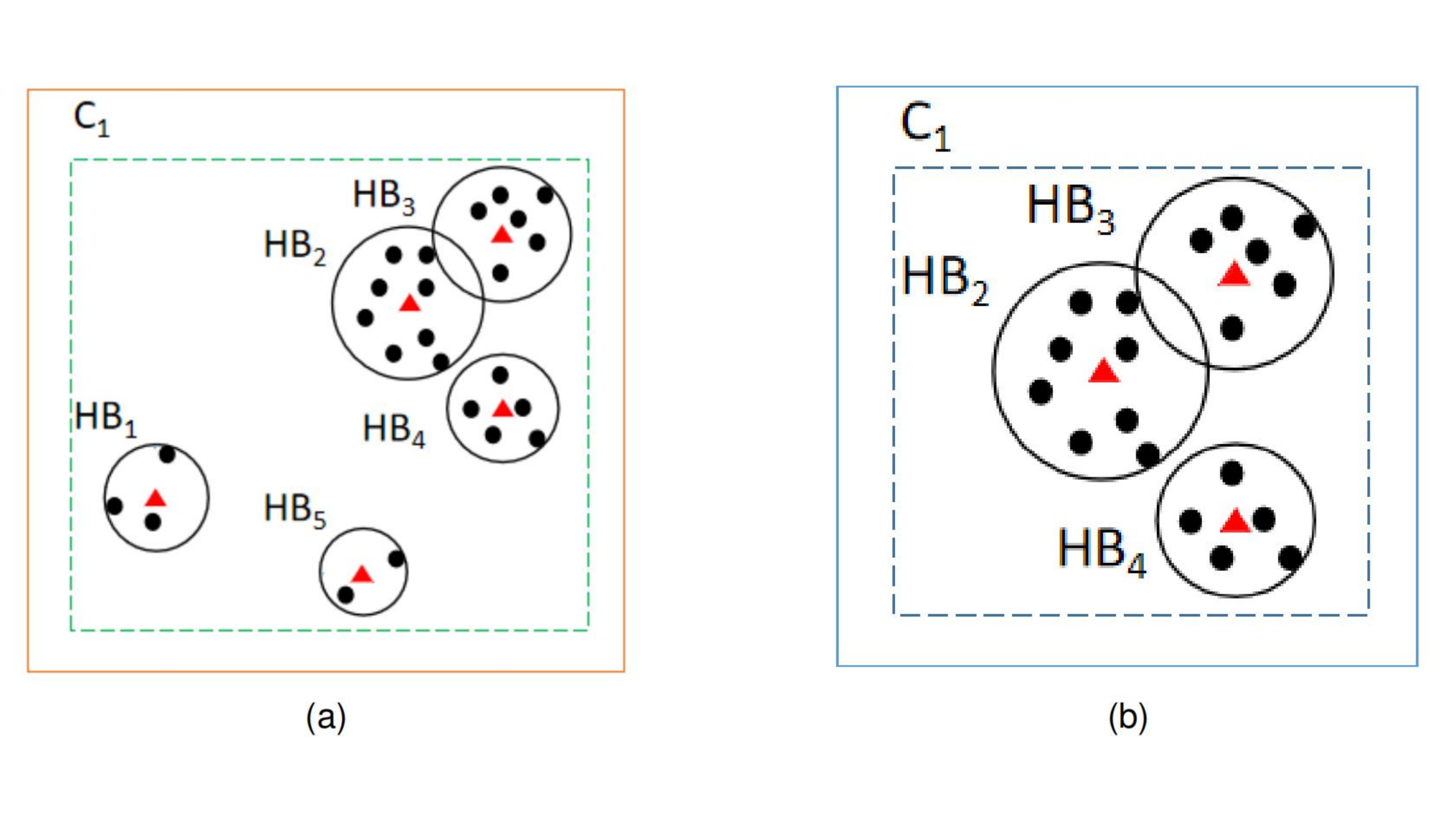}}
	\caption{(a) shows a data set $C_1$ containing noises with two noisy hyper-balls $C_{11}$ and $C_{15}$ which its samples is less than four. (b) shows the data set $C_1$ with denoised hyper-ball.}
	\label{fig_sim2}
\end{figure}

To increase the robustness of the HCVI index to the datasets containing noises, we propose an efficient method to deal with noisy datasets. As shown in Fig.3(a), $C_{1}$ represents one of the clusters in which there are five hyper-balls $HB_{1}$, $HB_{2}$, $HB_{3}$, $HB_{4}$ and $HB_{5}$. We regard it as a noise hyper-ball and remove it when the number of points in the hyper-ball is less than 4 samples as an empirical value to eliminate noise, such as the two hyper-balls $HB_{1}$ and $HB_{5}$, and Fig.3(b) shows the data set with denoised hyper-ball. Then the index is performed according to the method shown in Fig.2 calculation.
\section{Definitions of HCVI}
\textbf{Definition 3(intracluster compactness).} The intracluster compactness com ($C_k$), namely the value with the maximum distance between all two hyper-balls in the cluster, is defined as follows:
\begin{equation}
\label{deqn_ex5a}
com(C_k) =max(dist(HB_i,HB_j)).
\end{equation}

Where $HB_{i}\in{C_{k}}$, $HB_{j}\in{C_{k}}$ and When two hyper-balls belong to the overlapping relationship or there is only one hyper-ball in a cluster, the value is replaced with the minimum distance between all non-overlapping hyper balls within a cluster in all clusters as the distance of overlapping hyper balls.

\textbf{Definition 4(intercluster separation).} The intercluster separation sep($C_k$), namely the minimum value of minimum distances between all hyper-ball in the current cluster and all hyper-balls in other clusters, where $HB_i\in{C_k}$, $HB_j\in{C_u}(u\neq{k})$, is defined as follows:
\begin{equation}
\label{deqn_ex7a}
sep(C_k) =min(dist(HB_i,HB_{j})).
\end{equation}

When two hyper-balls $HB_{i}$ and $HB_{j}$ belong to the overlapping relationship, the value is replaced with the smallest distance among all pairs of non-overlapping hyper-balls in different clusters.

\textbf{Definition 5(HCVI).} Employing the ratio of intercluster separation to compactness, HCVI is computed as follows:
\begin{equation}
\label{deqn_ex9a}
HCVI(C_k)=com(C_k)/sep(C_k).
\end{equation}

The definition of clustering results which are evaluated by the average HCVI value of all clusters, where m is the number of hyper-ball generated and l is the number of clusters, is defined as follows:
\begin{equation}
\label{deqn_ex10a}
avgHCVI(l)=\mathop{1/l}\limits_{2<=l<=\sqrt{m}}\sum_{k=1}^{l}HCVI(C_k).
\end{equation}

Finally, we normalize the HCVI index through divide by the the largest value in avgHCVI(l).

\begin{quote}
{\bf References and Notes}

\end{quote}

\newpage
\vspace{11pt}
\vspace{-33pt}
\begin{wrapfigure}{l}{25mm}
	\includegraphics [width=1in,height=1.25in,clip,keepaspectratio]{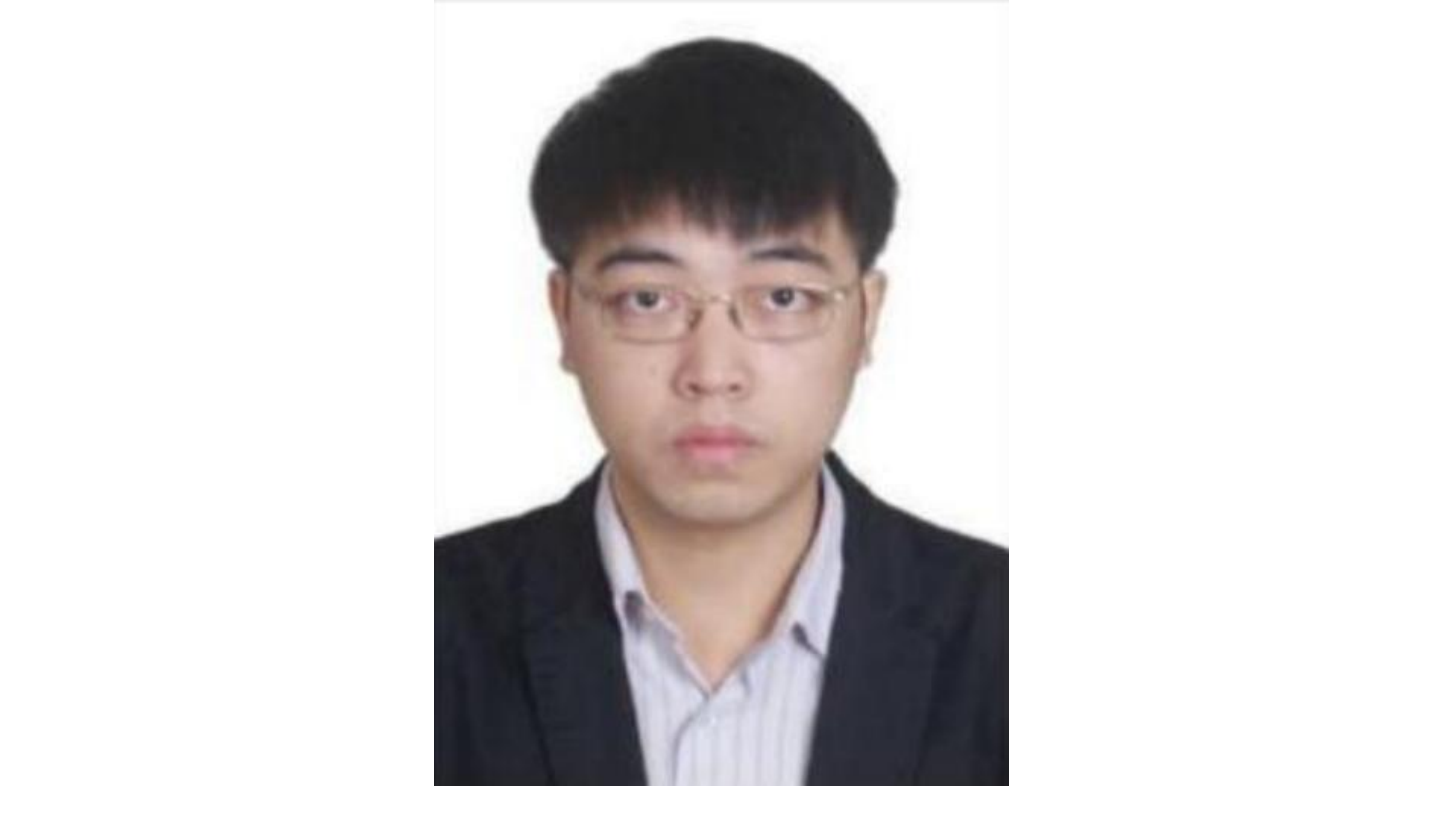}
\end{wrapfigure}\par
\textbf{Jiang Xie} received the MS and PhD degrees in computer science from Chongqing University, in 2015 and 2019, respectively. He is now a lecturer in the College of Computer Science and Technology at Chongqing University of Posts and Telecommunications. His research interests include clustering analysis and data mining.\par
\begin{wrapfigure}{l}{25mm}
	\includegraphics [width=1in,height=1.25in,clip,keepaspectratio]{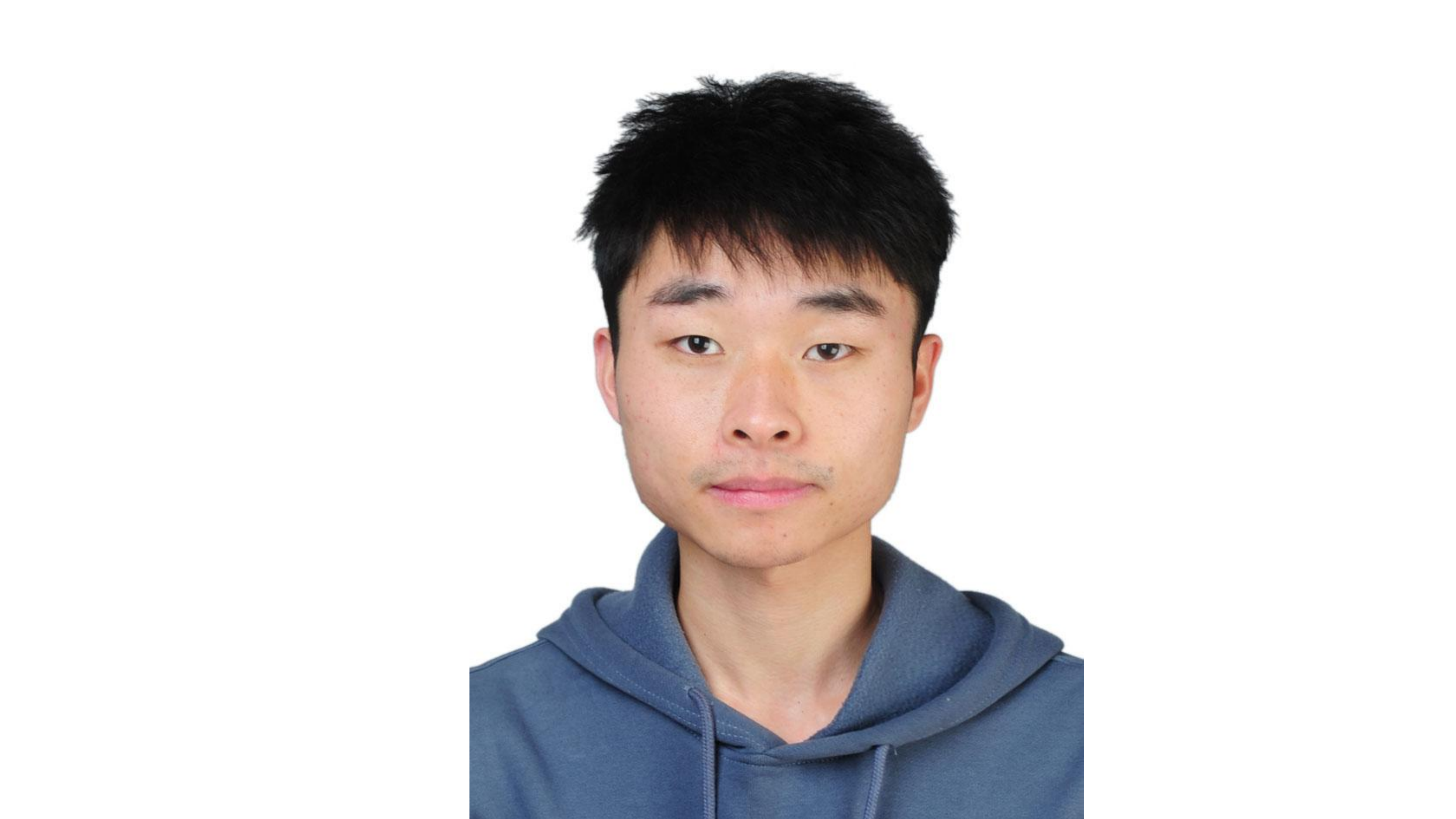}
\end{wrapfigure}\par
\textbf{Pengfei Zhao} received his bachelor's degree in 2017 from Anhui University of Science and Technology of China. He is currently a graduate student at College of Computer Science and Technology, Chongqing University of Posts and Telecommunications. His research interests include clustering analysis and data mining.\par
\begin{wrapfigure}{l}{25mm}
	\includegraphics [width=1in,height=1.25in,clip,keepaspectratio]{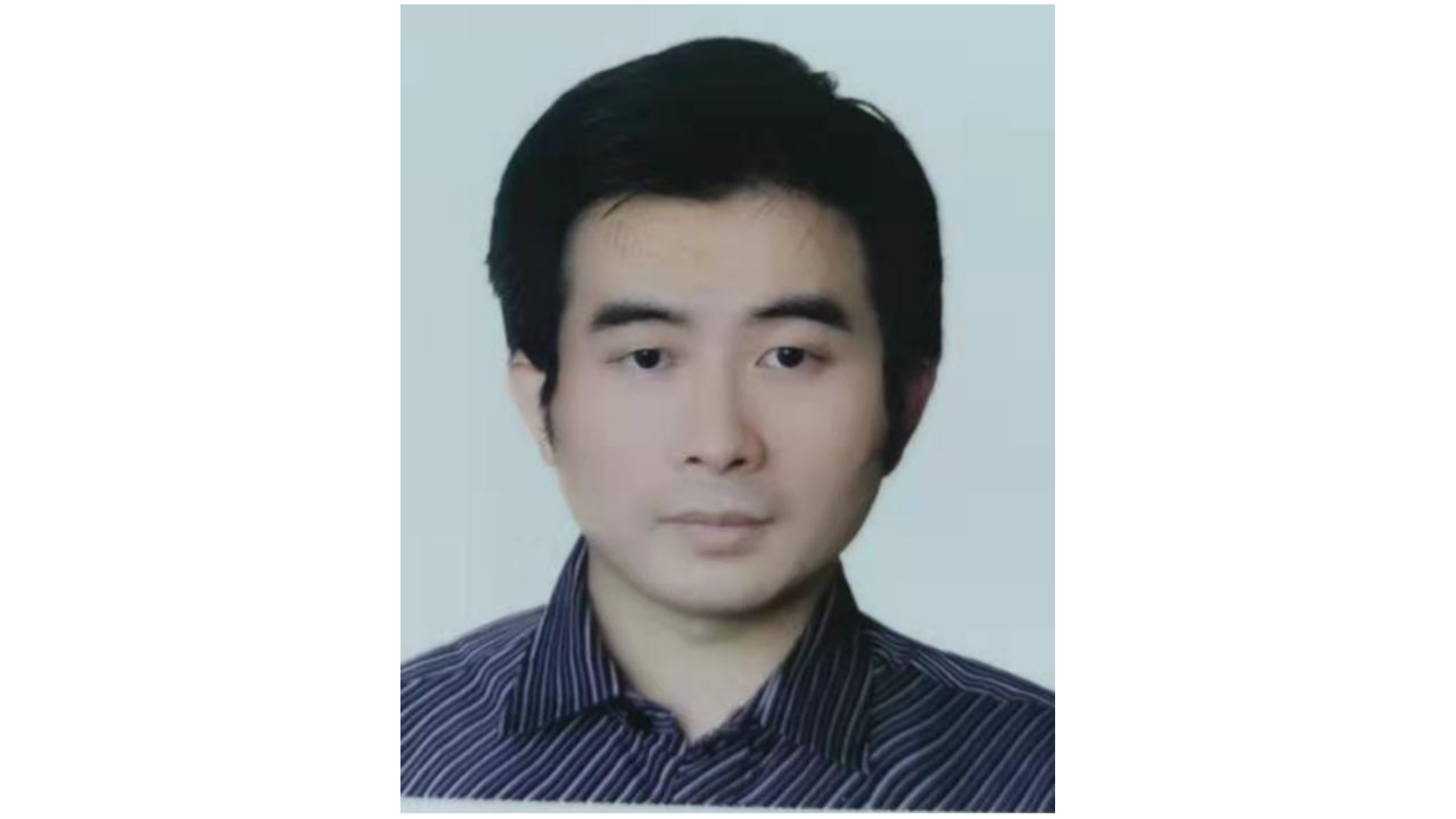}
\end{wrapfigure}\par
\textbf{Shuyin Xia*} received his B.S. degree and M.S. degree in Computer science in 2008 and 2012, respectively, both from Chongqing University of Technology in China. He received his Ph.D. degree from College of Computer Science in Chongqing University in China. He is an IEEE Member. Since 2015, he has been working at the Chongqing University of Posts and Telecommunications, Chongqing, China, where he is currently an associate professor and a Ph.D. supervisor, the executive deputy director of CQUPT - Chongqing Municipal Public Security Bureau - Qihoo 360 Big Data and Network Security Joint Lab. Dr. Xia is the director of Chongqing Artificial Intelligence Association. His research results have expounded at many prestigious journals, such IEEE-TKDE and IS. His research interests include data mining, granular computing, fuzzy rough sets, classifiers and label noise detection.\par
\begin{wrapfigure}{l}{25mm}
	\includegraphics [width=1in,height=1.25in,clip,keepaspectratio]{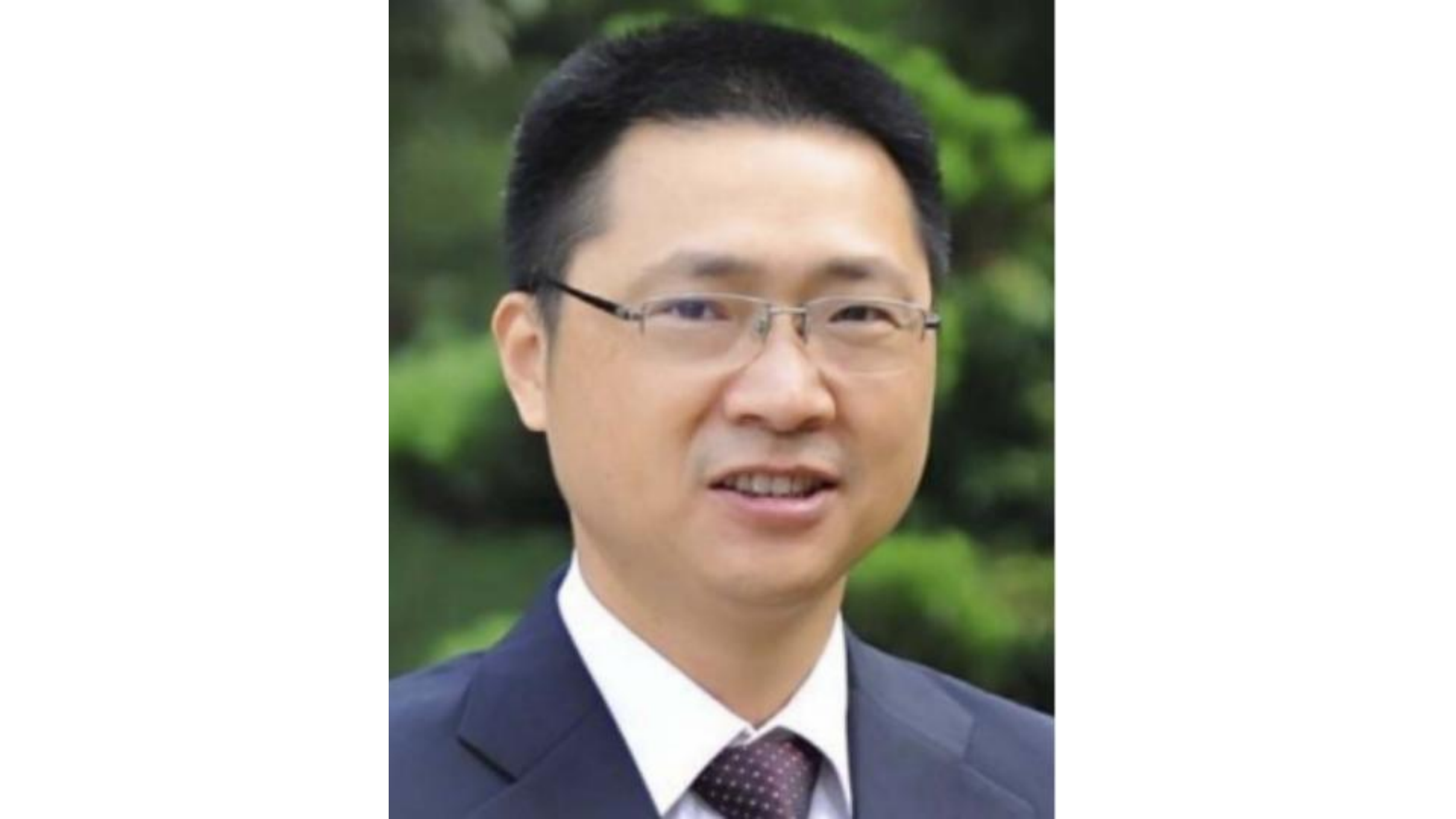}
\end{wrapfigure}\par
\textbf{Guoyin Wang(Senior Member, IEEE)} received a B.E. degree in computer software in 1992, a M.S. degree in computer software in 1994, and a Ph.D. degree in computer organization and architecture in 1996, all from Xi’an Jiaotong University in Xi’an, China. His research interests include data mining, machine learning, rough sets, granular computing, cognitive computing, and so forth. He has published over 300 papers in prestigious journals and conferences, including IEEE T-PAMI, T-KDE, T-IP, T-NNLS, and T-CYB. He has worked at the University of North T exas, USA, and the University of Regina, Canada, as a Visiting Scholar. Since 1996, he has been working at the Chongqing University of Posts and Telecommunications in Chongqing, China, where he is currently a Professor and a Ph.D. supervisor, the Director of the Chongqing Key Laboratory of Computational Intelligence, and the Vice President of the Chongqing University of Posts and Telecommunications. He is the Steering Committee Chair of the International Rough Set Society (IRSS), a Vice-President of the Chinese Association for Artificial Intelligence (CAAI), and a council member of the China Computer Federation (CCF).\par
\begin{wrapfigure}{l}{25mm}
	\includegraphics [width=1in,height=1.25in,clip,keepaspectratio]{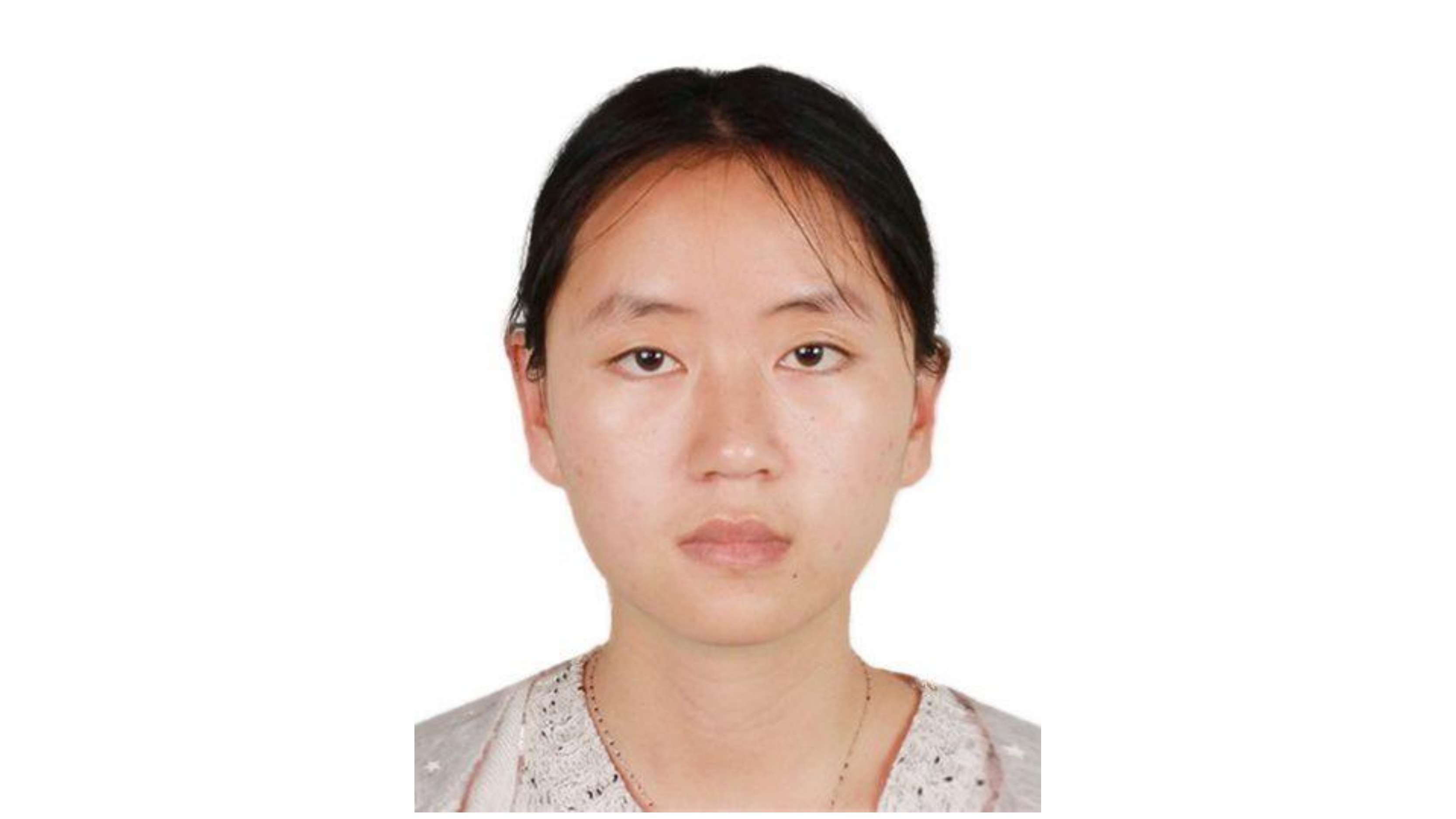}
\end{wrapfigure}\par
\textbf{Dongdong Cheng} received the bachelor’s degree in computer science from Chongqing Normal University, in 2013, and the doctor’s degree from Chongqing University, in 2018. She is now a lecturer in the College of Big Data and Intelligent Engineering at Yangtze Normal University. Her research interests include clustering analysis and data mining.\par

\vfill

\end{document}